\documentclass[12pt, letterpaper]{article}
\usepackage{graphicx}
\usepackage[margin=.75in]{geometry} 
\graphicspath{{.}}
\usepackage{authblk}  
\usepackage{hyperref} 
\usepackage{orcidlink} 
\usepackage[font=small]{caption} 
\usepackage{microtype}
\usepackage{float}
\usepackage[nodayofweek]{datetime}

\title{\textbf{\LARGE Interactive Classification Metrics: A graphical application to build robust intuition for classification model evaluation}}

\author[1]{David H. Brown\orcidlink{0000-0002-0969-8711}\thanks{Corresponding author. Email: \href{mailto:icm.python.app@gmail.com}{icm.python.app@gmail.com}}}
\author[2,3]{Davide Chicco\orcidlink{0000-0001-9655-7142}}

\affil[1]{\small Independent Researcher, Austin, Texas, USA}
\affil[2]{\small Dipartimento di Informatica Sistemistica e Comunicazione, Università di Milano-Bicocca, Milan, Italy}
\affil[3]{\small Institute of Health Policy Management and Evaluation, University of Toronto, Toronto, Ontario, Canada
\newline
\newline

Article version: \today}

\date{} 

\begin{document}

\maketitle

\begin{abstract}
    Machine learning continues to grow in popularity in academia, in industry, and is increasingly used in other fields. However, most of the common metrics used to evaluate even simple binary classification models have shortcomings that are neither immediately obvious nor consistently taught to practitioners. Here we present Interactive Classification Metrics (ICM), an application to visualize and explore the relationships between different evaluation metrics. The user changes the distribution statistics and explores corresponding changes across a suite of evaluation metrics. The interactive, graphical nature of this tool emphasizes the tradeoffs of each metric without the overhead of data wrangling and model training. The goals of this application are: (1) to aid practitioners in the ever-expanding machine learning field to choose the most appropriate evaluation metrics for their classification problem; (2) to promote careful attention to interpretation that is required even in the simplest scenarios like binary classification. Our application is publicly available for free under the MIT license as a Python package on PyPI at \href{https://pypi.org/project/interactive-classification-metrics}{https://pypi.org/project/interactive-classification-metrics} and on GitHub at \href{https://github.com/davhbrown/interactive_classification_metrics}{https://github.com/davhbrown/interactive\_classification\_metrics}. 

\end{abstract}

\textbf{Keywords:} machine learning, classification, Matthews Correlation Coefficient, ROC curve

\section{Statement of Need}
    More people enter the machine learning field every year through both formal and self-taught paths. In the United States alone, employment as a “Data Scientist” is expected to outpace the average over the next decade, with similar growth in other countries \cite{BLS:2024, Kaggle:2022}. User-friendly, low barrier to entry libraries like scikit-learn continue to be used by 80\% of modelers surveyed across the industry internationally \cite{Kaggle:2022}. Fields beyond statistics increasingly use machine learning tools, but are not always familiar with basic best practices like out-of-fold validation and avoiding data leakage, not to mention careful interpretation of evaluation metrics \cite{Kaggle:2022, Liu:2019, Kapoor:2023, Checkroud:2024}.
    
    Classification models are a fundamental tool in machine learning. The quality of a classification model is evaluated by comparing model predictions with ground truth targets, forming sections of the confusion matrix, resulting in the True Positive Rate, True Negative Rate, Positive Predictive Value, and Negative Predictive Value (NPV). Further evaluation metrics have been derived from these four “basic rates” \cite{Chicco:2023}, each summarizing different aspects of a model’s predictive performance, and each with advantages and disadvantages. For the simplest case of binary classification (two categories), the most common metrics and evaluation plots (Receiver Operating Characteristic (ROC) curve, the area under it (AUC), and Precision-Recall curve) have specific caveats known to experts, but not immediately apparent to novices, and can even be overlooked by experienced modelers. For example, their interpretation depends on class (im)balance and comparison to each other. This nuance only expands with modern multi-class classification (for example, evaluation of semantic segmentation models in computer vision). While these metrics do not claim to capture all four quadrants of the confusion matrix, in practice they are often reported on their own or as a single number that adequately describes a model’s predictive quality.

    Literature details the shortcomings with ROC curves \cite{Chicco:2023} and singular metrics \cite{Powers:2020}; has evaluated less common metrics like bookmaker informedness and Matthews Correlation Coefficient (MCC; \cite{Chicco:2023, Chicco:2021}); and even proposed novel graphical tools like the MCC-F1 curve \cite{Cao:2020}. However, insights from scholarly work take time to enter widespread educational material, and are not immediately obvious from static examples or mathematical formulae alone.

    It is essential that the machine learning community have every pedagogical resource at its disposal to solidify a deep understanding of and intuition for the field’s most basic, but nuanced, model evaluation tools. ICM introduced here aims to improve the choices novice scientists make, and deepen the intuition that experienced practitioners have for evaluating classification models.

\section{Overview of Functionality}
    After installing ICM in a Python environment, it is callable from the terminal with a single command, which opens a web browser for interaction. It runs on the user’s local machine using bokeh server. Underlying computations are handled by common Python libraries for data/machine learning.

\begin{figure}[ht]
        \centering
        \includegraphics[width=0.9\textwidth]{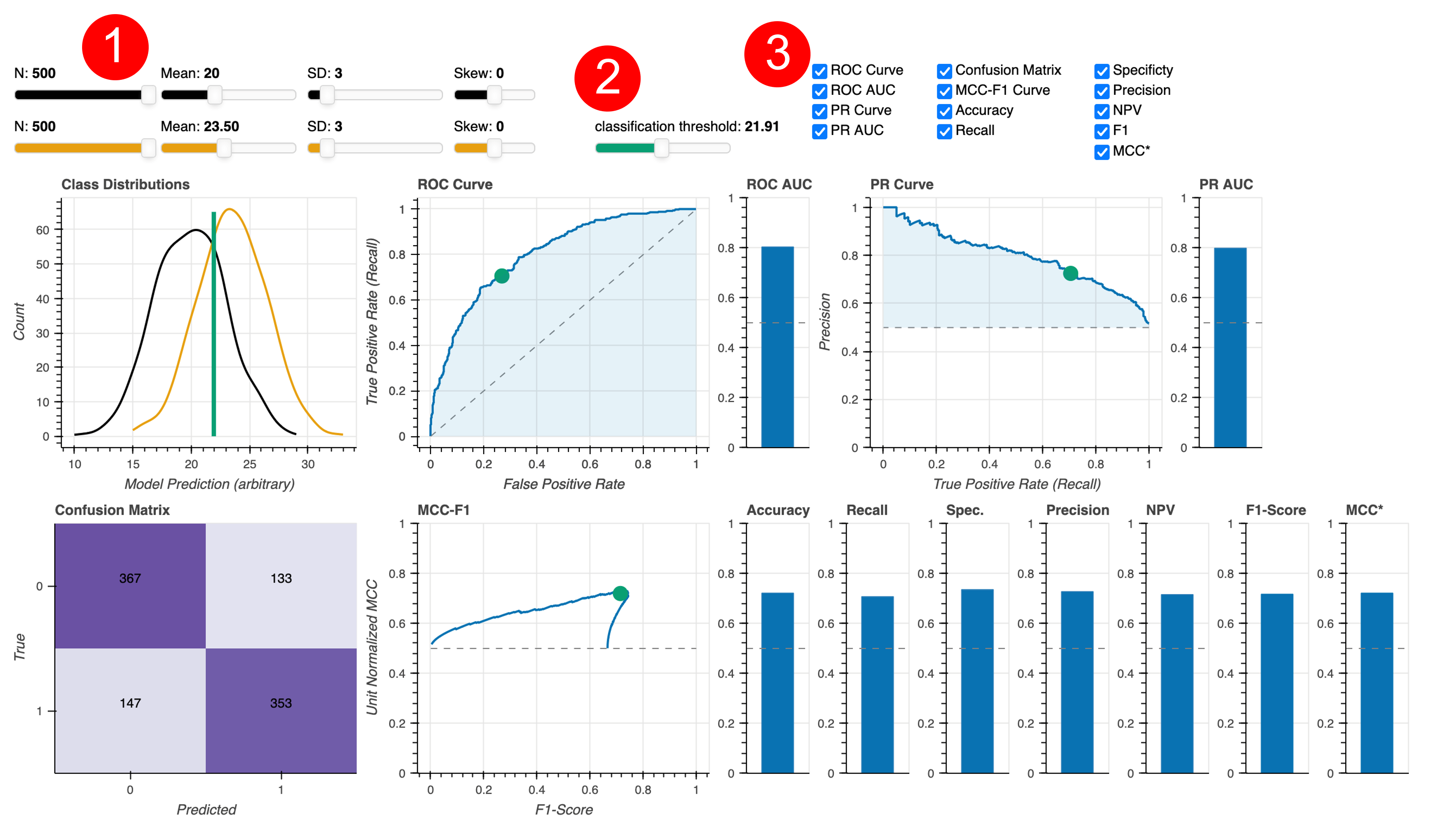}
        \captionsetup{width=0.9\textwidth}
        \caption{Screenshot of the application with numbered steps overlaid (red circles). Users control 9 interactive sliders at the top, and all graphs respond accordingly. The sliders control the sample size (N), mean, standard deviation (SD), and skew of the two distributions (\textbf{Step 1}) that represent the negative (black) and positive class \emph{predictions} (orange). The properties of these distributions, along with the classification threshold (green; \textbf{Step 2}) control the magnitude and shape of all other plots. Users can also choose to show or hide specific plots with checkboxes (\textbf{Step 3}). Full display shown.}
        \label{fig:Figure1}
    \end{figure}

    \textbf{\autoref{fig:Figure1}} shows the application; a brief animation of the application in use is available in the GitHub repository. The full list of plots, evaluation metrics, and their acronyms are: Class distributions\footnote{Note that the Class Distributions do not represent different distributions of input features. The correct interpretation of the Class Distributions is a model’s predictions, the output, with one change in the horizontal axis range: the predictions do not range from 0.0-1.0 as they often would for class probabilities. This is purely a user interface design choice that allows the horizontal axis to be unbounded as the user adjusts the mean and SD sliders. The overall lesson of this application is that the relative size and shape of the two distributions affects the outcome of all other evaluation metrics, without specific attention to their absolute values. Specialists can consider these horizontal axis values similar to the output of a neural network before the softmax function is applied.}, ROC curve, Area Under the ROC curve (ROC AUC), Precision-Recall (PR) curve, Area Under the PR curve (PR AUC), Confusion Matrix, Matthews Correlation Coefficient-F1 (MCC-F1) curve, Accuracy, Recall, Specificity, Precision, Negative Predictive Value (NPV), F1-Score, and MCC (unit normalized). Different fields have different naming conventions for the same metric; for example, Recall is also called the True Positive Rate, Sensitivity, or Hit Rate. Wikipedia maintains a comprehensive list of alternative naming conventions for these metrics \cite{website:wiki-confusion-matrix}.

\section{Example Scenario}   
\begin{figure}[H]
        \centering
        \includegraphics[width=0.9\textwidth]{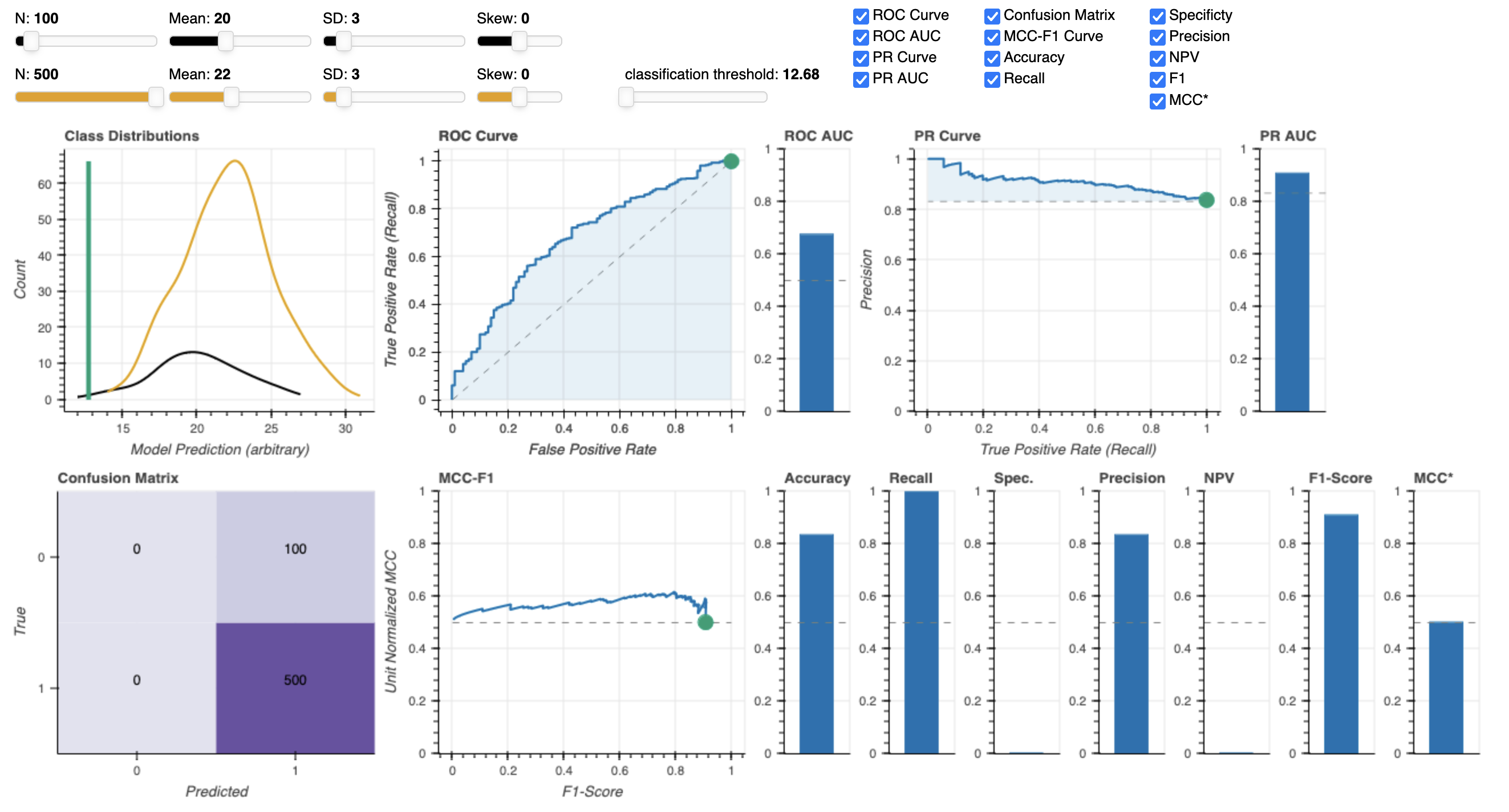}
        \captionsetup{width=0.9\textwidth}
        \caption{The classic flaw of Accuracy on an imbalanced dataset. The negative class (black) has N=100 examples, the positive class (orange) has N=500. The classification threshold (green) is set extremely low to represent a model that predicts everything as the positive class, yet achieves over 80\% Accuracy due to the proportions of the two classes in the dataset.}
        \label{fig:Figure2}
    \end{figure}

    \textbf{\autoref{fig:Figure2}} depicts ICM demonstrating a classic model evaluation mistake: using Accuracy alone to evaluate an overfit model trained on an imbalanced dataset. A comprehensive look at all available metrics reveals the flaw: with additional information beyond Accuracy and ROC AUC, it is clear that when accounting for all four “basic rates”, MCC reveals chance performance (MCC = 0.5). Also note that PR AUC is 0.9, which could be naively interpreted as a good model, but its minimum baseline is 0.83, driven by the proportion of classes. This emphasizes that the area covered by the PR curve is quite small, and that PR AUC out of context is misleading. The shading under the PR curve, which only shades above the baseline, makes this point immediately obvious in graphical form.
    
\section{Prior Work}
    At the time of writing, we know of no formal visualization tool or similar application that displays all of these metrics in a dynamic, interactive format with the specific purpose of manipulating distributions to understand the pros and cons of different evaluation metrics. Good resources have been published in other online venues, but are either static, or not as comprehensive in terms of the number of evaluation metrics they provide \cite{website:Wilber-ROC, website:Wilber-PR, image:wiki-ROC, website:navan-ROC, website:google-ROC}.
    
    Numerous textbooks and courses use Accuracy along with imbalanced classes as a classical example of the importance of careful interpretation of metrics, but at the time of writing we have not found popular course material or textbooks that cover the additional issues of ROC/PR-curve tradeoffs, the strengths of MCC, and the newer MCC-F1 curve as well as the original literature does. This application enables quick demonstration of these relationships easily for a wide audience, either during formal instruction or self-driven exploration.

\section{Conclusion}
    Visualizing results is a pivotal task in many fields, and scientific visualization is a cornerstone of modern data science \cite{Midway:2020}. Software tools can make this task easier for researchers and students, but to the best of our knowledge, no Python package for visualizing binary classification results in an interactive, exploratory way exists. We fill this gap by providing our ICM software. Our application does not require the overhead of cleaning data and training models. Instead, it allows users to focus on understanding the evaluation metrics themselves, and how these metrics change in relation to each other as the user alters the statistics of the two distributions.

\section*{Additional Sections}

\paragraph{List of abbreviations}
AUC:~Area Under the Curve.
ICM:~Interactive Classification Metrics.
MCC:~Matthews Correlation Coefficient.
NPV:~Negative Predictive Value.
PPV:~Positive Predictive Value, Precision.
PR:~Precision-Recall.
ROC:~Receiver Operating Characteristic.
TNR:~True Negative Rate, Specificity.
TPR:~True Positive Rate, Recall, Sensitivity.

\paragraph{Acknowledgments}
The authors acknowledge Arthur Colombini Gusmão, the author of \texttt{py-mcc-f1} used in this application.

\paragraph{Funding}
The work of D.C. is partially funded by the Italian Ministero Italiano delle Imprese e del Made in Italy under the Digital Intervention in Psychiatric and Psychologist Services~(DIPPS) programme and is partially supported by Ministero dell’Università e della Ricerca of Italy under the ``Dipartimenti di Eccellenza 2023-2027'' ReGAInS grant assigned to Dipartimento di Informatica Sistemistica e Comunicazione at Università di Milano-Bicocca. The funders had no role in study design, data collection and analysis, decision to publish, or preparation of the manuscript.

\paragraph{Conflict of interest}
The authors declare no conflict of interest in this work. At the time of writing DHB is employed by an artificial intelligence company. The software described in this paper was developed independently by DHB as a personal project. His employer is not affiliated with the development of the software, has no ownership rights, and did not provide any resources or support for this work. His contributions to the project were made entirely outside of his employment with any entity.

\paragraph{Software availability}
Our software code is publicly available for free under the MIT license as a Python package on PyPI at \href{https://pypi.org/project/interactive-classification-metrics}{https://pypi.org/project/interactive-classification-metrics} 
\\
and on GitHub at \href{https://github.com/davhbrown/interactive_classification_metrics}{https://github.com/davhbrown/interactive\_classification\_metrics}

\paragraph{Ethical committee and data consent for collection}
Not applicabile.

\paragraph{Author contributions}
{DHB}: conceptualization, software implementation, application design, writing. {DC}: application design, domain expertise, scholarly integration, writing

\bibliographystyle{IEEEtran}
\bibliography{refs}

\begin{thebibliography}{10}
\providecommand{\url}[1]{#1}
\csname url@samestyle\endcsname
\providecommand{\newblock}{\relax}
\providecommand{\bibinfo}[2]{#2}
\providecommand{\BIBentrySTDinterwordspacing}{\spaceskip=0pt\relax}
\providecommand{\BIBentryALTinterwordstretchfactor}{4}
\providecommand{\BIBentryALTinterwordspacing}{\spaceskip=\fontdimen2\font plus
\BIBentryALTinterwordstretchfactor\fontdimen3\font minus \fontdimen4\font\relax}
\providecommand{\BIBforeignlanguage}[2]{{%
\expandafter\ifx\csname l@#1\endcsname\relax
\typeout{** WARNING: IEEEtran.bst: No hyphenation pattern has been}%
\typeout{** loaded for the language `#1'. Using the pattern for}%
\typeout{** the default language instead.}%
\else
\language=\csname l@#1\endcsname
\fi
#2}}
\providecommand{\BIBdecl}{\relax}
\BIBdecl

\bibitem{BLS:2024}
\BIBentryALTinterwordspacing
{Bureau of Labor Statistics, U.S. Department of Labor}, ``Occupational outlook handbook, data scientists,'' 2024. [Online]. Available: \url{https://www.bls.gov/ooh/math/data-scientists.htm}
\BIBentrySTDinterwordspacing

\bibitem{Kaggle:2022}
\BIBentryALTinterwordspacing
Kaggle, ``State of data science and machine learning,'' 2022. [Online]. Available: \url{https://www.kaggle.com/kaggle-survey-2022}
\BIBentrySTDinterwordspacing

\bibitem{Liu:2019}
\BIBentryALTinterwordspacing
X.~Liu, L.~Faes, A.~U. Kale, S.~K. Wagner, D.~J. Fu, A.~Bruynseels, T.~Mahendiran, G.~Moraes, M.~Shamdas, C.~Kern, J.~R. Ledsam, M.~K. Schmid, K.~Balaskas, E.~J. Topol, L.~M. Bachmann, P.~A. Keane, and A.~K. Denniston, ``A comparison of deep learning performance against health-care professionals in detecting diseases from medical imaging: a systematic review and meta-analysis,'' \emph{The Lancet Digital Health}, vol.~1, no.~6, pp. e271--e297, 2019. [Online]. Available: \url{https://doi.org/10.1016/S2589-7500(19)30123-2}
\BIBentrySTDinterwordspacing

\bibitem{Kapoor:2023}
\BIBentryALTinterwordspacing
S.~Kapoor and A.~Narayanan, ``Leakage and the reproducibility crisis in machine-learning-based science,'' \emph{Patterns}, vol.~4, no.~9, p. 100804, 2023. [Online]. Available: \url{https://doi.org/10.1016/j.patter.2023.100804}
\BIBentrySTDinterwordspacing

\bibitem{Checkroud:2024}
\BIBentryALTinterwordspacing
A.~M. Chekroud, M.~Hawrilenko, H.~Loho, J.~Bondar, R.~Gueorguieva, A.~Hasan, J.~Kambeitz, P.~R. Corlett, N.~Koutsouleris, H.~M. Krumholz, J.~H. Krystal, and M.~Paulus, ``Illusory generalizability of clinical prediction models,'' \emph{Science}, vol. 383, no. 6679, pp. 164--167, 2024. [Online]. Available: \url{https://www.science.org/doi/abs/10.1126/science.adg8538}
\BIBentrySTDinterwordspacing

\bibitem{Chicco:2023}
\BIBentryALTinterwordspacing
D.~Chicco and G.~Jurman, ``The {M}atthews correlation coefficient ({MCC}) should replace the {ROC AUC} as the standard metric for assessing binary classification,'' \emph{BioData Mining}, vol.~16, no.~4, pp. 1--23, 2023. [Online]. Available: \url{https://doi.org/10.1186/s13040-023-00322-4}
\BIBentrySTDinterwordspacing

\bibitem{Powers:2020}
\BIBentryALTinterwordspacing
D.~M.~W. Powers, ``Evaluation: from precision, recall and {F-measure to ROC}, informedness, markedness and correlation,'' 2020. [Online]. Available: \url{https://arxiv.org/abs/2010.16061}
\BIBentrySTDinterwordspacing

\bibitem{Chicco:2021}
\BIBentryALTinterwordspacing
D.~Chicco, N.~Tötsch, and G.~Jurman, ``The {M}atthews correlation coefficient ({MCC}) is more reliable than balanced accuracy, bookmaker informedness, and markedness in two-class confusion matrix evaluation,'' \emph{BioData Mining}, vol.~14, pp. 1--22, 2021. [Online]. Available: \url{https://doi.org/10.1186/s13040-021-00244-z}
\BIBentrySTDinterwordspacing

\bibitem{Cao:2020}
\BIBentryALTinterwordspacing
C.~Cao, D.~Chicco, and M.~M. Hoffman, ``The mcc-f1 curve: a performance evaluation technique for binary classification,'' 2020. [Online]. Available: \url{https://arxiv.org/abs/2006.11278}
\BIBentrySTDinterwordspacing

\bibitem{website:wiki-confusion-matrix}
\BIBentryALTinterwordspacing
Wikipedia, ``Table of confusion.'' [Online]. Available: \url{https://en.wikipedia.org/wiki/Confusion_matrix#Table_of_confusion}
\BIBentrySTDinterwordspacing

\bibitem{website:Wilber-ROC}
\BIBentryALTinterwordspacing
J.~Wilber, ``{ROC and AUC}: A visual explanation of receiver operating characteristic curves and area under the curve,'' June 2022. [Online]. Available: \url{https://mlu-explain.github.io/roc-auc/}
\BIBentrySTDinterwordspacing

\bibitem{website:Wilber-PR}
\BIBentryALTinterwordspacing
------, ``Precision and recall: Accuracy is not enough,'' March 2022. [Online]. Available: \url{https://mlu-explain.github.io/precision-recall/}
\BIBentrySTDinterwordspacing

\bibitem{image:wiki-ROC}
\BIBentryALTinterwordspacing
Sharpr and Kakau, ``{ROC} curves,'' October 2015. [Online]. Available: \url{https://upload.wikimedia.org/wikipedia/commons/4/4f/ROC_curves.svg}
\BIBentrySTDinterwordspacing

\bibitem{website:navan-ROC}
\BIBentryALTinterwordspacing
R.~Navaneethakrishnan, ``Understanding {ROC} curves,'' October 2014. [Online]. Available: \url{http://navan.name/roc/}
\BIBentrySTDinterwordspacing

\bibitem{website:google-ROC}
\BIBentryALTinterwordspacing
Google, ``Classification: Accuracy, recall, precision, and related metrics,'' November 2024. [Online]. Available: \url{https://developers.google.com/machine-learning/crash-course/classification/accuracy-precision-recall}
\BIBentrySTDinterwordspacing

\bibitem{Midway:2020}
\BIBentryALTinterwordspacing
S.~R. Midway, ``Principles of effective data visualization,'' \emph{Patterns}, vol.~1, no.~9, 2020. [Online]. Available: \url{https://doi.org/10.1016/j.patter.2020.100141}
\BIBentrySTDinterwordspacing

\end{thebibliography}

\end{document}